\documentclass[runningheads]{llncs}

\usepackage{hyperref}
\usepackage{graphicx}
\usepackage{multirow}
\usepackage{amssymb}
\usepackage{amsmath}
\usepackage{booktabs}

\begin{document}
\title{Leveraging Labeling Representations in Uncertainty-based Semi-supervised Segmentation}

%
\titlerunning{Leveraging Labeling Representations in Uncertainty-based SSL}


\author{Sukesh Adiga V \thanks{Corresponding author:~{sukesh.adiga-vasudeva.1@etsmtl.net}}  \and Jose Dolz \and Herve Lombaert}

\authorrunning{Adiga et al.}
\institute{ETS Montreal, Canada}

%

\maketitle              
\begin{abstract}

Semi-supervised segmentation tackles the scarcity of annotations by leveraging unlabeled data with a small amount of labeled data. A prominent way to utilize the unlabeled data is by consistency training which commonly uses a teacher-student network, where a teacher guides a student segmentation.
The predictions of unlabeled data are not reliable, therefore, uncertainty-aware methods have been proposed to gradually learn from meaningful and reliable predictions. Uncertainty estimation, however, relies on multiple inferences from model predictions that need to be computed for each training step, which is computationally expensive. 
This work proposes a novel method to estimate the pixel-level uncertainty by leveraging the labeling representation of segmentation masks. On the one hand, a labeling representation is learnt to represent the available segmentation masks. The learnt labeling representation is used to map the prediction of the segmentation into a set of plausible masks. Such a reconstructed segmentation mask aids in estimating the pixel-level uncertainty guiding the segmentation network.
The proposed method estimates the uncertainty with a single inference from the labeling representation, thereby reducing the total computation. We evaluate our method on the 3D segmentation of left atrium in MRI, and we show that our uncertainty estimates from our labeling representation improve the segmentation accuracy over state-of-the-art methods.

\keywords{Semi-Supervised learning \and Segmentation \and Labeling Representation \and Uncertainty.}

\end{abstract}

\section{Introduction}

Segmentation of organs or abnormal regions is a fundamental task in clinical applications, such as diagnosis, intervention and treatment planning. Deep learning techniques are driving progress in automating the segmentation task under the full-supervision paradigm \cite{milletari2016v,cciccek20163d}. Training these models, however, relies on a large amount of pixel-level annotations, which require expensive clinical expertise \cite{cheplygina2019not}.

Semi-supervised learning (SSL) techniques alleviate the annotation scarcity by leveraging unlabeled data with a small amount of labeled data. Current semi-supervised segmentation methods typically utilize the unlabeled data either in the form of pseudo labels \cite{bai2017semi,zheng2020cartilage}, regularization \cite{nie2018asdnet,cui2019semi,peng2020deep} or knowledge priors \cite{zheng2019semi,he2020dense}. For instance, self-training methods \cite{bai2017semi} generate pseudo labels from unlabeled data, which are used to retrain the network iteratively. A wide range of regularization-based methods has been explored for semi-supervised segmentation using adversarial learning \cite{nie2018asdnet,chaitanya2019semi}, consistency learning \cite{bortsova2019semi,yu2019uncertainty,li2020transformation,luo2021semi}, or co-training \cite{peng2020deep,xia20203d,wang2021self}. Adversarial methods encourage the segmentation of unlabeled images to be closer to those of the labeled images. In contrast, consistency and co-training methods encourage two or more segmentation predictions, either from the same or different networks, to be consistent under different perturbations of the input data. Such consistency-based methods are popular in semi-supervision due to their simplicity. Consequently, self-ensembling \cite{laine2016temporal} and mean teacher-based \cite{tarvainen2017mean} methods are often used in semi-supervised segmentation of medical images \cite{cui2019semi,bortsova2019semi,li2020transformation}. However, their generated predictions from the unlabeled images may not always be reliable. To alleviate this issue, uncertainty-aware regularization methods \cite{yu2019uncertainty,sedai2019uncertainty,wang2020double,wang2021tripled,luo2021efficient} are proposed to gradually add reliable target regions in predictions. This uncertainty scheme is also employed in co-training \cite{xia20203d} and self-training \cite{zheng2020cartilage} approaches to obtain reliable predictions. Although these methods perform well in low-labeled data regimes, their high computation and complex training techniques might limit their applicability to broader applications in practice. For instance, the uncertainty estimation is approximated via Monte-Carlo Dropout \cite{gal2016dropout} or an ensembling, which requires multiple predictions per image. Co-training methods require two or more networks to be trained simultaneously, whereas self-training-based methods rely on costly iterations. Lastly, adversarial training is challenging in terms of convergence \cite{salimans2016improved}.

Prior-based methods in semi-supervised segmentation typically incorporate anatomical knowledge of the target object during training the model. For instance, He \textit{et al.} \cite{he2020dense} encode the unlabeled images in an autoencoder and combine the learnt features as prior knowledge in the segmentation networks. Recent attempts use signed distance maps (SDM) as shape constraints during training \cite{li2020shape,xue2020shape,wang2021tripled}. For instance, Le \textit{et al.} \cite{li2020shape} propose an additional task of predicting SDM and enforcing consistency with an adversarial loss. Zheng \textit{et al.} \cite{zheng2019semi} exploit a probabilistic atlas in their loss function. These knowledge-based methods require an additional task to constraints shape prior, or it requires aligned images.

These limitations motivate our approach, which leverages a learnt labeling representation to approximate the uncertainty. Our main idea is to mimic a shape prior by learning a representation using segmentation masks such that each prediction is mapped into a set of plausible segmentations. In contrast to \cite{zheng2019semi}, our approach does not require aligned images. The mapped segmentation is subsequently used to estimate the uncertainty maps to guide the segmentation network.
We hypothesize that the proposed uncertainty estimates are more robust than those derived from the entropy variance, requiring multiple inferences strategy.

\paragraph{\bf Our contributions.}

We propose a novel way to estimate the pixel-wise uncertainty to guide the training of a segmentation model. In particular, we integrate a pre-trained denoising autoencoder (DAE) into the training, whose goal is to leverage a learnt labeling representation on unlabeled data. The DAE maps the segmentation predictions into a set of plausible segmentation masks. Then, we approximate the uncertainty by computing the pixel-wise difference between predicted segmentation and its DAE reconstruction. In contrast to commonly used uncertainty-based approaches, our uncertainty map needs a single inference from the DAE model, reducing computation complexity. Our method is extensively evaluated on the 2018 Atrial Segmentation Challenge dataset \cite{xiong2021global}. The results demonstrate the superiority of our approach over the state-of-the-art.

\begin{figure*}[t!]
\centering
\includegraphics[width=0.975\linewidth]{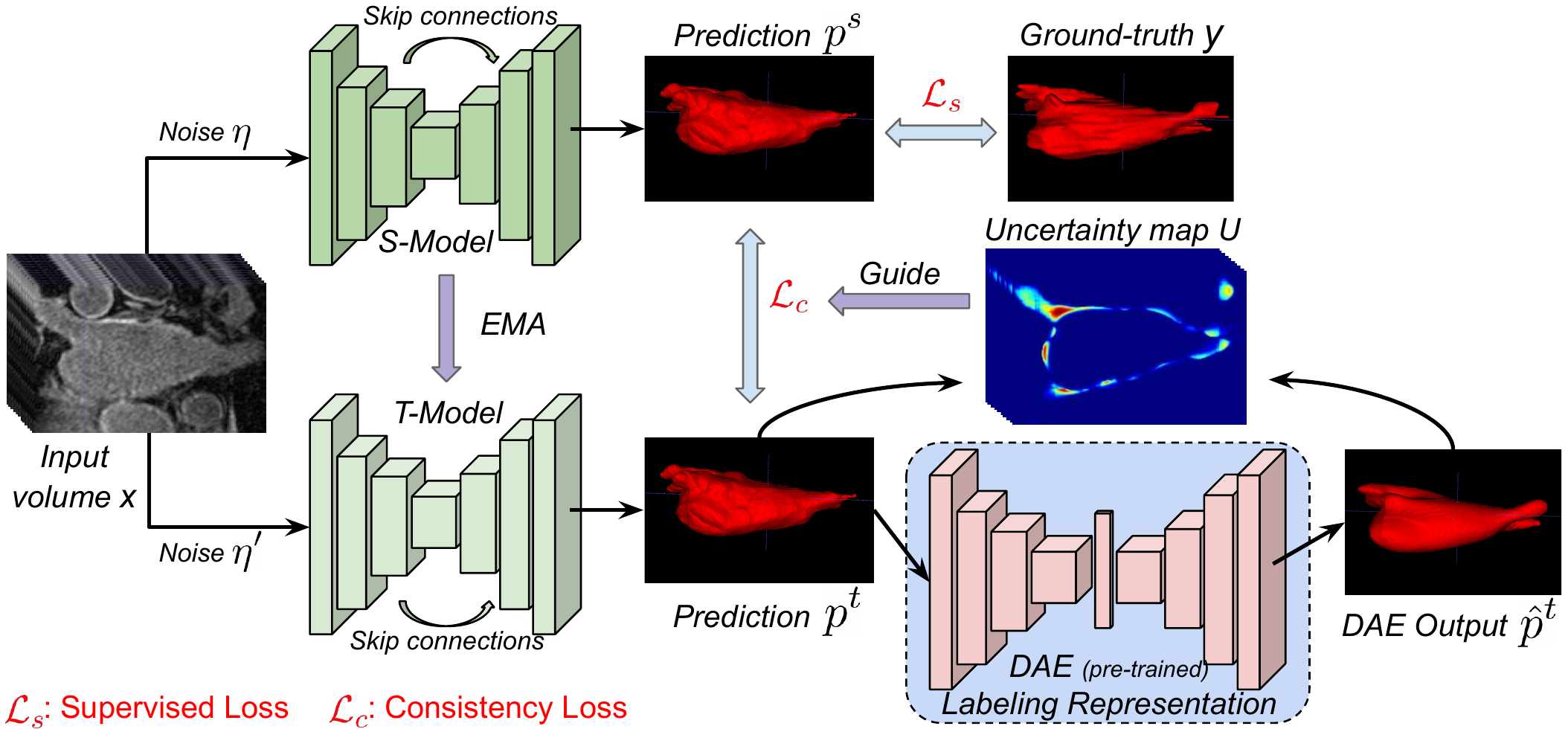}
\caption{Overview of our uncertainty estimation from labeling representation for semi-supervised segmentation. A pre-trained labeling representation (DAE) is integrated into the training of the mean teacher method, which maps the teacher predictions $p^t$ into plausible segmentation $\hat{p}^t$. The uncertainty map ($U$) is subsequently estimated with the teacher and DAE predictions, guiding the student model.}
\label{fig:arch}
\end{figure*}

\section{Method}
\label{sec:methods}

The schematic of the proposed label representation-based uncertainty estimation is shown in Fig~\ref{fig:arch}. The main idea is to exploit a labeling representation that maps the predictions of the segmentation into set of plausible masks. The reconstructed segmentations will be later employed to estimate an uncertainty map. Following current literature \cite{yu2019uncertainty}, we adopt a mean teacher approach to train a segmentation network. These steps are detailed next.

\subsection{Mean Teacher Formulation}
The standard semi-supervised learning consists of $N$ labeled and $M$ unlabeled data in the training set, where $N \ll M$. Let  $D_L = \{(x_i, y_i)\}_{i=1}^N$ and $D_U = \{(x_i)\}_{N+1}^{(N+M)}$ denote the labeled and unlabeled sets, where an input volume is represented as $x_i \in R^{H \times W \times D}$ and its corresponding segmentation mask is $y_i \in \{0,1,...,C\}^{H \times W \times D}$, with $C$ being the number of classes. We use the common mean teacher approach used in semi-supervised segmentation, which consists of a student ($S$) and teacher ($T$) model, both having the same segmentation architecture. The overall objective function is defined as follows:

\begin{equation}
\label{eq:learner}
\mathcal{L} = \underset{\theta_s}{\text{min}} \sum_{i=1}^N \mathcal{L}_s(f(x_i; \theta_s), y_i) + \lambda_c \sum_{i=1}^{N+M} \mathcal{L}_c(f(x_i; \theta_s, \eta), f(x_i; \theta_t; \eta{'} )),
\end{equation}

where $f(\cdot)$ denotes the segmentation network, and $\theta_s$ and $\theta_t$ are the learnable weights of the student and teacher models. The supervised loss $\mathcal{L}_s$ measures the segmentation quality on the labeled data, whereas the consistency loss $\mathcal{L}_c$ measures the prediction consistency of student and teacher models for the same input volume $x_i$ under different perturbations ($\eta$, $\eta{'}$). 
The balance between supervised and unsupervised loss is controlled by a ramp-up weighting co-efficient $\lambda_c$. In the mean teacher training, the student model parameters are optimized with stochastic gradient descent (SGD), whereas exponential moving average (EMA) is employed at each training step $t$, i.e., $\theta_t = \alpha \theta_{t-1} + (1-\alpha) \theta_s$ to update the teacher model parameters. Note that $\alpha$ is the smoothing coefficient of EMA that controls the update rate.

\subsection{Labeling Representation Prior}
Incorporating object shape prior in deep segmentation models is not obvious. One of the reasons is that, in order to integrate such prior knowledge during training, one needs to augment the learning objective with a differentiable term, which in the case of complex shapes is not trivial. To circumvent these difficulties, a simpler solution is to resort to an autoencoder trained with pixel-wise labels, which can represent anatomical priors and be used as a global regularizer during training. This strategy has been adopted for fully supervised training in \cite{oktay2017anatomically} and as a post-processing step in \cite{larrazabal2020post} to correct the segmentation predictions. Motivated by this, we represent the available labels in a non-linear latent space using a denoising autoencoder (DAE) \cite{vincent2010stacked}, which somehow mimics a shape prior. The DAE model consists of an encoder $f_e(\cdot)$ and a decoder module $f_d(\cdot)$ with a $d$-dimensional latent space as shown in the Fig.~\ref{fig:arch}. The DAE is trained to reconstruct the clean labels $y_i$ from its corrupted version $\tilde{y}_i$, which can be achieved with a mean squared error loss: $\frac{1}{H \times W \times D}\sum_v ||f_d(f_e(\tilde{y}_{i,v})) - y_{i,v}||^2$.

\subsection{Uncertainty from a Labeling Representation}
The role of the uncertainty is to gradually update the student model with reliable target regions from the teacher predictions. 
Our proposed method estimates the uncertainty directly from the labeling representation network $f_d(f_e(\cdot))$, 
requiring only one inference step. First, we map 
the prediction from the teacher model $p^t_i$ with a DAE model to produce a plausible segmentation $\hat{p}^t_i$. We subsequently estimate the uncertainty as the pixel-wise difference between the DAE output and the prediction, i.e., $U_i = ||\hat{p}^t_i - p^t_i||^2$. Then, the reliable target for the consistency loss is obtained as $e^{-\gamma U_i}$, similarly to \cite{luo2021efficient}, where $\gamma$ is an uncertainty weighting factor empirically set to 1. Finally, our consistency loss is defined as: 

\begin{equation}
\mathcal{L}_c(p^s_i, p^t_i) = \frac{\sum_v e^{-\gamma U_{i,v}} ||p^s_{i,v} - p^t_{i,v}||^2}{\sum_v e^{-\gamma U_{i,v}}}
\end{equation}

where $v$ is a voxel. We jointly optimize the consistency loss $\mathcal{L}_c$ and supervised loss $\mathcal{L}_s$ as learning objectives, where $\mathcal{L}_s$ uses the cross-entropy and dice losses.

\section{Results}
Our proposed method is compared with the state-of-the-art semi-supervised segmentation methods \cite{yu2019uncertainty,li2020shape,luo2021semi,luo2021efficient} \footnote{\small{We use official implementation provided by baseline methods to run the experiments.}}. We group the uncertainty-based methods to assess the effectiveness of our uncertainty estimation for segmentation. For a fair comparison, all experiments are run three times with a fixed set of seeds on the same machines, and their average results are reported.

\paragraph{\bf Dataset and Evaluation Metrics.}
Our method is evaluated on the Left Atrium (LA) dataset from the 2018 Atrial Segmentation Challenge \cite{xiong2021global}. The dataset consists of 100 3D MR volumes of LA with an isotropic resolution of 0.625$mm^3$ and corresponding segmentation masks. In our experiments, we use a 80/20 training/testing split and apply the same preprocessing as in \cite{yu2019uncertainty,li2020shape,luo2021semi}. The training set is partitioned into $N/M$ labeled/unlabeled splits, fixed across all methods for each setting. We employ Dice Score Coefficient (DSC) and 95\% Hausdorff Distance (HD) metrics to assess segmentation performance.    


\paragraph{\bf Implementation and Training details.}
Following \cite{yu2019uncertainty,li2020shape,luo2021semi}, we use V-net \cite{milletari2016v} as backbone architecture for teacher, student and DAE models. The skip connections are removed, and a dense layer is added at the bottleneck layer for the DAE model. The student model is trained by a SGD optimizer with an initial learning rate ($lr$) of 0.1 and momentum 0.9 for 6000 iterations with a cosine annealing \cite{loshchilov2016sgdr} decaying. The teacher weights are updated by an EMA with an update rate of $\alpha$=0.99 as in \cite{tarvainen2017mean}. The consistency weight is updated with Gaussian warming up function $\lambda_c = \beta * e^{-5(1-t/t_{max})^2}$, where $t$ and $t_{max}$ denotes current and maximum training iterations, and $\beta$ is set to 0.1, as in \cite{yu2019uncertainty}. The DAE model is also trained with SGD with $lr$=0.1, momentum of 0.9 and decaying the $lr$ by 2 for every 5000 iterations. Input to both segmentation and DAE networks are random cropped to $112 \times 112 \times 80$ size and employ online standard data augmentation techniques such as random flipping and rotation. In addition, input labels to the DAE model are corrupted with a random swapping of pixels around class boundaries, morphological operations (erosion and dilation), resizing and adding/removing shapes. The batch size is set to 4 in both networks. Input batch for segmentation network uses two labeled and unlabeled data. For testing, generating segmentation predictions uses the sliding window strategy, and the method is evaluated at the last iteration as in \cite{yu2019uncertainty}. Our experiments were run on an NVIDIA RTX A6000 GPU with PyTorch 1.8.0+cu111.

\paragraph{\bf Comparison with the state-of-the-art.}
We now compare our method with relevant semi-supervised segmentation approaches under the 10\% and 20\% labeled data settings and report their results in Tables~\ref{table:10results}-\ref{table:20results}. Non-uncertainty-based method such as MT \cite{tarvainen2017mean}, DCT \cite{luo2021semi}, and SASSnet \cite{li2020shape} are grouped in the middle of the table, while uncertainty-based methods UAMT \cite{yu2019uncertainty}, URPC \cite{luo2021efficient} \footnote{Note that URPC \cite{luo2021efficient} use multi-scale 3D U-Net \cite{cciccek20163d} architecture.} and our methods are grouped at the bottom of each table. The upper and lower bound from the backbone architecture V-net \cite{milletari2016v} are reported in the top.

In the first setting, 10\% of labeled data is used, and the remaining images are used as unlabeled data. From Table~\ref{table:10results}, we can observe that leveraging unlabeled data improves the lower bound in all baselines. The uncertainty-based baselines seem to improve the segmentation performance by 1\% in Dice score compared to non-uncertainty-based baselines. However, their performance drops in terms of HD up to 5mm. Among baseline methods, UAMT and DCT achieve the best Dice and HD scores, respectively. Compared to these best performing baselines, our method brings 1.5\% and 0.8mm improvements in Dice and HD scores. Moreover, uncertainty estimation in our method requires a single inference from a labeling representation, whereas UAMT uses $K$=8 inferences per training step to obtain an uncertainty map.

Furthermore, we also validate our method on the 20\% of labeled data scenario, whose results are reported in Tables~\ref{table:20results}. Results demonstrate a similar trend as compared to the 10\% experiments. The uncertainty-based baselines improve 1\% in terms of Dice and drop up to 1mm in HD, compared to non-uncertainty-based methods. Our method improves the best performing baseline in both Dice and HD scores. Particularly, our method improves the HD score by 2.5mm compared to the best performing baseline (SASSnet).

\begin{figure*}[t!]
\centering
\includegraphics[width=0.9\linewidth]{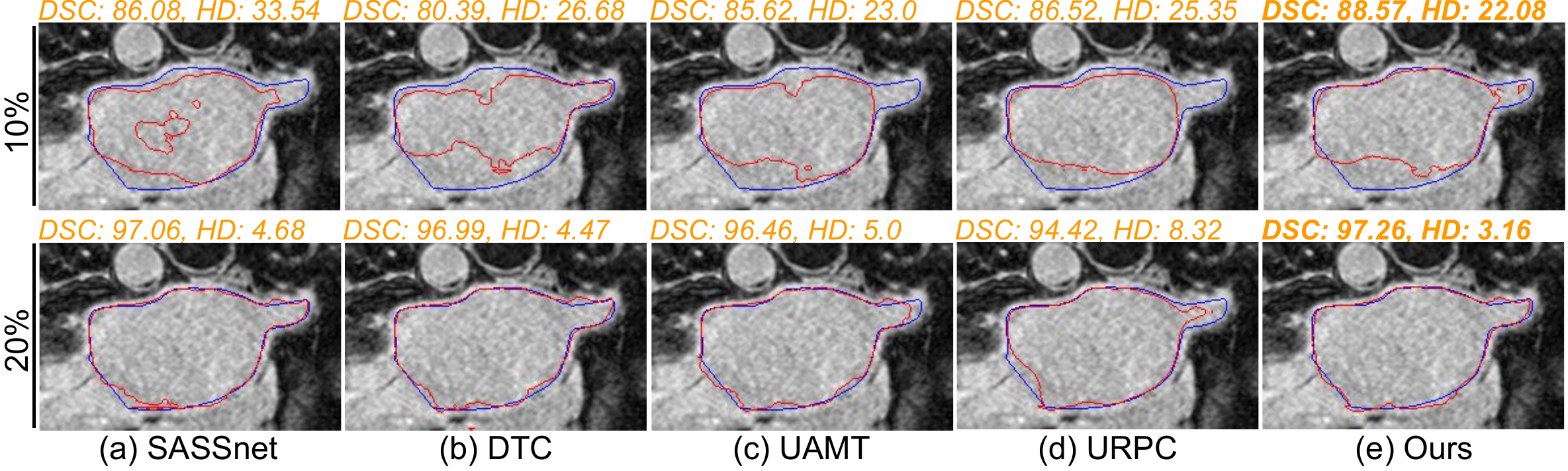}
\caption{\textbf{Qualitative comparison under 10\% and 20\% annotation setting.} DSC (\%) and HD (mm) scores are mentioned at the top of each image. Coloring is  the prediction (Red) and ground truth (Blue).}
\label{fig:seg_results}
\end{figure*}

Visual results of different segmentation results are depicted in Fig.~\ref{fig:seg_results}. In the top row of the figure, the segmentation of SASSnet produces holes in segmentation, and their method employs a post-processing tool to improve the segmentation, which is avoided for a fair comparison. DTC captures the challenging top right side region in segmentation; however, the prediction is under-segmented and noisy. The uncertainty-based methods improve the segmentation in UAMT and produce smooth segmentation boundaries in URPC. Our method improves the segmentation region further compared to URPC. In the case of 20\% labeled data experiments, all methods improve the segmentation due to having access to more labels during training, while the boundary regions are either under or over-segmented. Our method produces better and smoother segmentation, which can be due to the knowledge derived from the labeling representation.

\begin{table}[h!]
\centering
\addtolength{\tabcolsep}{9pt}
\caption{Segmentation results on the LA test set for 10\% labeled data experiments averaged over three runs. Uncertainty methods with $K$ inferences are grouped at the bottom, while $K$ = -, indicates non-uncertainty methods.}
\scalebox{0.9}{
\begin{tabular}{l | c | c | c c}
\toprule
\bf Methods         & \bf \#K & \bf $N$/$M$  & \textbf{DSC (\%)} & \textbf{HD (mm)} \\
\midrule
Upper bound      & - & 80/0  & 91.23 $\pm$ 0.44	& 6.08 $\pm$ 1.84    \\
Lower bound      & - & 8/0   & 76.07 $\pm$ 5.02	& 28.75 $\pm$ 0.72    \\

\midrule
MT \cite{tarvainen2017mean}     & - & 8/72 & 78.22 $\pm$ 6.89	& 16.74 $\pm$ 4.80	\\
SASSnet \cite{li2020shape}      & - & 8/72 & 83.70 $\pm$ 1.48	& 16.90 $\pm$ 1.35	\\
DCT \cite{luo2021semi}          & - & 8/72 & 83.10 $\pm$ 0.26	& 12.62 $\pm$ 1.44	\\

\midrule
UAMT \cite{yu2019uncertainty}   & 8 & 8/72 & 85.09 $\pm$ 1.42	& 18.34 $\pm$ 2.80	\\
URPC \cite{luo2021efficient}    & 1 & 8/72 & 84.47 $\pm$ 0.31	& 17.11 $\pm$ 0.60	\\

Ours        & 1 & 8/72  & \bf 86.58 $\pm$ 1.03	& \bf 11.82 $\pm$ 1.42	\\
\bottomrule

\end{tabular}}
\label{table:10results}
\end{table}

\begin{table}[h!]
\centering
\addtolength{\tabcolsep}{9pt}
\caption{Segmentation results on the LA test set for 20\% labeled data experiments averaged over three runs. Uncertainty methods with $K$ inferences are grouped at the bottom, while $K$ = -, indicates non-uncertainty methods.}
\scalebox{0.9}{
\begin{tabular}{l | c | c | c c}
\toprule
\bf Methods         & \bf \#K & \bf $N$/$M$  & \textbf{DSC (\%)} & \textbf{HD (mm)}\\
\midrule
Upper bound     & - & 80/0  & 91.23 $\pm$ 0.44	& 6.08 $\pm$ 1.84   \\
Lower bound     & - & 16/0  & 81.46 $\pm$ 2.96	& 23.61 $\pm$ 4.94  \\

\midrule
MT \cite{tarvainen2017mean}     & - & 16/64 & 86.06 $\pm$ 0.81	& 11.63 $\pm$ 3.4  \\
SASSnet \cite{li2020shape}      & - & 16/64 & 87.81 $\pm$ 1.45	& 10.18 $\pm$ 0.55	\\
DCT \cite{luo2021semi}          & - & 16/64 & 87.35 $\pm$ 1.26	& 10.25 $\pm$ 2.49	\\

\midrule
UAMT \cite{yu2019uncertainty}   & 8 & 16/64 & 87.78 $\pm$ 1.03	& 11.1 $\pm$ 1.91  \\
URPC \cite{luo2021efficient}    & 1 & 16/64 & 88.58 $\pm$ 0.10	& 13.1 $\pm$ 0.60	\\

Ours        & 1 & 16/64  & \bf 88.60 $\pm$ 0.82	& \bf 7.61 $\pm$ 0.78   \\
\bottomrule

\end{tabular}}
\label{table:20results}
\end{table}

\paragraph{\bf Ablation Study.}
To validate the effectiveness of our uncertainty estimation on segmentation performance, two experiments are conducted by adopting threshold strategy and entropy scheme from UAMT. Particularly, a threshold strategy is used in consistency, whereas entropy is used to estimate the uncertainty, and their results are reported in Table~\ref{table:ablation}. Compared to UAMT, our threshold and entropy experiments significantly improve the segmentation performance in HD and Dice scores, while our proposed method (L2-based exponential uncertainty) achieves the best performance. These results show the merit of our labeling representation for uncertainty estimation. Furthermore, we report the ablation on uncertainty weight $\gamma$ and consistency weight $\beta$, in Table~\ref{table:ablation_beta_gamma}. Results demonstrate that $\gamma$=1 is best for our method, while for $\beta$=1 our method further improves Dice and HD scores; however, we report on $\beta$=0.1 in all experiments for a fair comparison. Overall, for most of the $\gamma$ and $\beta$ values, our method is consistently better than UAMT baselines, demonstrating the robustness of our approach.

\begin{table}[h!]
\centering
\caption{Effectiveness of our proposed uncertainty estimation on segmentation results.}
\scalebox{0.9}{
\begin{tabular}{l | c | c c}
\toprule
\bf Methods         &  \bf $N$/$M$  & \textbf{DSC (\%)} & \textbf{HD (mm)} \\

\midrule
UAMT \cite{yu2019uncertainty}   & 8/72  & 85.09 $\pm$ 1.42	& 18.34 $\pm$ 2.80	\\
Ours (Threshold)                & 8/72  & 85.39 $\pm$ 0.91	& 12.96 $\pm$ 3.05 \\
Ours (Entropy)                  & 8/72  & 85.92 $\pm$ 1.52	& \bf 11.16 $\pm$ 0.82 \\
Ours                            & 8/72  & \bf 86.58 $\pm$ 1.03	& 11.82 $\pm$ 1.42 \\
\bottomrule

\end{tabular}}
\label{table:ablation}
\end{table}

\begin{table}[h!]
\centering
\addtolength{\tabcolsep}{2.0pt}
\caption{Evaluating the $\gamma$ and $\beta$ values under 10\% annotation setting.}
\scalebox{0.9}{
\begin{tabular}{l c c | l c c}
\toprule
\bf $\gamma$, $\beta=0.1$ & \textbf{DSC (\%)} & \textbf{HD (mm)} & \bf $\beta$, $\gamma=1$ & \textbf{DSC (\%)} & \textbf{HD (mm)} \\

\midrule
0.1 & 85.30 $\pm$ 1.17      & 13.51 $\pm$ 2.66      & 0.01  & 84.89 $\pm$ 0.92      & 11.84 $\pm$ 2.79      \\
0.5 & 85.28 $\pm$ 0.60      & 14.01 $\pm$ 4.44      & 0.05  & 85.88 $\pm$ 1.44      & 10.98 $\pm$ 1.85      \\
1   & \bf 86.58 $\pm$ 1.03  & \bf 11.82 $\pm$ 1.42  & 0.1   & 86.58 $\pm$ 1.03	    & 11.82 $\pm$ 1.42      \\
2   & 85.84 $\pm$ 1.39	    & 12.13 $\pm$ 3.43      & 0.5   & 86.54 $\pm$ 0.74	    & 12.42 $\pm$ 1.31      \\
5   & 84.87 $\pm$ 0.85	    & 15.28 $\pm$ 1.76      & 1     & \bf 86.89 $\pm$ 0.6   & \bf 9.85 $\pm$ 0.82   \\
\bottomrule

\end{tabular}}
\label{table:ablation_beta_gamma}
\end{table}

\section{Conclusion}
We presented a novel labeling representation-based uncertainty estimation for the semi-supervised segmentation. Our method produces an uncertainty map from a labeling representation network, which guides the reliable regions of prediction for the segmentation network, thereby achieving better segmentation results. Results demonstrate that the proposed method achieves the best performance compared to state-of-the-art baselines on left atrium segmentation from 3D MR volumes in two different settings. 
The ablation studies demonstrate the effectiveness and robustness of our uncertainty estimation compared to the entropy-based method. Our proposed uncertainty estimation from the labeling representation approach can be adapted to a broader range of applications where it is crucial to obtain a reliable prediction.

\paragraph{\bf Acknowledgments:}
This research work was partly funded by the Canada Research Chair on Shape Analysis in Medical Imaging, the Natural Sciences and Engineering Research Council of Canada (NSERC), and the Fonds de Recherche du Quebec (FQRNT). 

\bibliographystyle{splncs04}
\bibliography{biblio}

\begin{thebibliography}{10}
\providecommand{\url}[1]{\texttt{#1}}
\providecommand{\urlprefix}{URL }
\providecommand{\doi}[1]{https://doi.org/#1}

\bibitem{bai2017semi}
Bai, W., Oktay, O., Sinclair, M., Suzuki, H., Rajchl, M., Tarroni, G., Glocker,
  B., King, A., Matthews, P.M., Rueckert, D.: Semi-supervised learning for
  network-based cardiac {MR} image segmentation. In: MICCAI. pp. 253--260.
  Springer (2017)

\bibitem{bortsova2019semi}
Bortsova, G., Dubost, F., Hogeweg, L., Katramados, I., Bruijne, M.d.:
  Semi-supervised medical image segmentation via learning consistency under
  transformations. In: MICCAI. pp. 810--818. Springer (2019)

\bibitem{chaitanya2019semi}
Chaitanya, K., Karani, N., Baumgartner, C.F., Becker, A., Donati, O.,
  Konukoglu, E.: Semi-supervised and task-driven data augmentation. In: IPMI.
  pp. 29--41. Springer (2019)

\bibitem{cheplygina2019not}
Cheplygina, V., de~Bruijne, M., Pluim, J.P.: Not-so-supervised: a survey of
  semi-supervised, multi-instance, and transfer learning in medical image
  analysis. MedIA  \textbf{54},  280--296 (2019)

\bibitem{cciccek20163d}
{\c{C}}i{\c{c}}ek, {\"O}., Abdulkadir, A., Lienkamp, S.S., Brox, T.,
  Ronneberger, O.: {3D U-Net}: learning dense volumetric segmentation from
  sparse annotation. In: MICCAI. pp. 424--432. Springer (2016)

\bibitem{cui2019semi}
Cui, W., Liu, Y., Li, Y., Guo, M., Li, Y., Li, X., Wang, T., Zeng, X., Ye, C.:
  Semi-supervised brain lesion segmentation with an adapted mean teacher model.
  In: IPMI. pp. 554--565. Springer (2019)

\bibitem{gal2016dropout}
Gal, Y., Ghahramani, Z.: Dropout as a bayesian approximation: Representing
  model uncertainty in deep learning. In: ICML. pp. 1050--1059. PMLR (2016)

\bibitem{he2020dense}
He, Y., Yang, G., Yang, J., Chen, Y., Kong, Y., Wu, J., Tang, L., Zhu, X.,
  Dillenseger, J.L., Shao, P., et~al.: Dense biased networks with deep priori
  anatomy and hard region adaptation: Semi-supervised learning for fine renal
  artery segmentation. MedIA  \textbf{63},  101722 (2020)

\bibitem{laine2016temporal}
Laine, S., Aila, T.: Temporal ensembling for semi-supervised learning. arXiv
  preprint arXiv:1610.02242  (2016)

\bibitem{larrazabal2020post}
Larrazabal, A.J., Mart{\'\i}nez, C., Glocker, B., Ferrante, E.: {Post-DAE:}
  anatomically plausible segmentation via post-processing with denoising
  autoencoders. IEEE TMI  \textbf{39}(12),  3813--3820 (2020)

\bibitem{li2020shape}
Li, S., Zhang, C., He, X.: Shape-aware semi-supervised {3D} semantic
  segmentation for medical images. In: MICCAI. pp. 552--561. Springer (2020)

\bibitem{li2020transformation}
Li, X., Yu, L., Chen, H., Fu, C.W., Xing, L., Heng, P.A.:
  Transformation-consistent self-ensembling model for semisupervised medical
  image segmentation. IEEE Transactions on Neural Networks and Learning Systems
   \textbf{32}(2),  523--534 (2020)

\bibitem{loshchilov2016sgdr}
Loshchilov, I., Hutter, F.: {SGDR:} stochastic gradient descent with warm
  restarts. arXiv preprint arXiv:1608.03983  (2016)

\bibitem{luo2021semi}
Luo, X., Chen, J., Song, T., Wang, G.: Semi-supervised medical image
  segmentation through dual-task consistency. In: AAAI. vol.~35, pp. 8801--8809
  (2021)

\bibitem{luo2021efficient}
Luo, X., Liao, W., Chen, J., Song, T., Chen, Y., Zhang, S., Chen, N., Wang, G.,
  Zhang, S.: Efficient semi-supervised gross target volume of nasopharyngeal
  carcinoma segmentation via uncertainty rectified pyramid consistency. In:
  MICCAI. pp. 318--329. Springer (2021)

\bibitem{milletari2016v}
Milletari, F., Navab, N., Ahmadi, S.A.: {V-net:} fully convolutional neural
  networks for volumetric medical image segmentation. In: 3DV. pp. 565--571.
  IEEE (2016)

\bibitem{nie2018asdnet}
Nie, D., Gao, Y., Wang, L., Shen, D.: {ASDNet:} attention based semi-supervised
  deep networks for medical image segmentation. In: MICCAI. pp. 370--378.
  Springer (2018)

\bibitem{oktay2017anatomically}
Oktay, O., Ferrante, E., Kamnitsas, K., Heinrich, M., Bai, W., Caballero, J.,
  Cook, S.A., De~Marvao, A., Dawes, T., O‘Regan, D.P., et~al.: Anatomically
  constrained neural networks {(ACNNs)}: application to cardiac image
  enhancement and segmentation. IEEE TMI  \textbf{37}(2),  384--395 (2017)

\bibitem{peng2020deep}
Peng, J., Estrada, G., Pedersoli, M., Desrosiers, C.: Deep co-training for
  semi-supervised image segmentation. Pattern Recognition  \textbf{107},
  107269 (2020)

\bibitem{salimans2016improved}
Salimans, T., Goodfellow, I., Zaremba, W., Cheung, V., Radford, A., Chen, X.:
  Improved techniques for training gans. NeurIPS  \textbf{29} (2016)

\bibitem{sedai2019uncertainty}
Sedai, S., Antony, B., Rai, R., Jones, K., Ishikawa, H., Schuman, J., Gadi, W.,
  Garnavi, R.: Uncertainty guided semi-supervised segmentation of retinal
  layers in {OCT} images. In: MICCAI. pp. 282--290. Springer (2019)

\bibitem{tarvainen2017mean}
Tarvainen, A., Valpola, H.: Mean teachers are better role models:
  Weight-averaged consistency targets improve semi-supervised deep learning
  results. NeurIPS  \textbf{30} (2017)

\bibitem{vincent2010stacked}
Vincent, P., Larochelle, H., Lajoie, I., Bengio, Y., Manzagol, P.A., Bottou,
  L.: Stacked denoising autoencoders: Learning useful representations in a deep
  network with a local denoising criterion. JMLR  \textbf{11}(12) (2010)

\bibitem{wang2021tripled}
Wang, K., Zhan, B., Zu, C., Wu, X., Zhou, J., Zhou, L., Wang, Y.:
  Tripled-uncertainty guided mean teacher model for semi-supervised medical
  image segmentation. In: MICCAI. pp. 450--460. Springer (2021)

\bibitem{wang2021self}
Wang, P., Peng, J., Pedersoli, M., Zhou, Y., Zhang, C., Desrosiers, C.:
  Self-paced and self-consistent co-training for semi-supervised image
  segmentation. MedIA  \textbf{73},  102146 (2021)

\bibitem{wang2020double}
Wang, Y., Zhang, Y., Tian, J., Zhong, C., Shi, Z., Zhang, Y., He, Z.:
  Double-uncertainty weighted method for semi-supervised learning. In: MICCAI.
  pp. 542--551. Springer (2020)

\bibitem{xia20203d}
Xia, Y., Liu, F., Yang, D., Cai, J., Yu, L., Zhu, Z., Xu, D., Yuille, A., Roth,
  H.: {3D} semi-supervised learning with uncertainty-aware multi-view
  co-training. In: IEEE/CVF WCCV. pp. 3646--3655 (2020)

\bibitem{xiong2021global}
Xiong, Z., Xia, Q., Hu, Z., Huang, N., Bian, C., Zheng, Y., Vesal, S.,
  Ravikumar, N., Maier, A., Yang, X., et~al.: A global benchmark of algorithms
  for segmenting the left atrium from late gadolinium-enhanced cardiac magnetic
  resonance imaging. MedIA  \textbf{67},  101832 (2021)

\bibitem{xue2020shape}
Xue, Y., Tang, H., Qiao, Z., Gong, G., Yin, Y., Qian, Z., Huang, C., Fan, W.,
  Huang, X.: Shape-aware organ segmentation by predicting signed distance maps.
  In: AAAI. vol.~34, pp. 12565--12572 (2020)

\bibitem{yu2019uncertainty}
Yu, L., Wang, S., Li, X., Fu, C.W., Heng, P.A.: Uncertainty-aware
  self-ensembling model for semi-supervised {3D} left atrium segmentation. In:
  MICCAI. pp. 605--613. Springer (2019)

\bibitem{zheng2019semi}
Zheng, H., Lin, L., Hu, H., Zhang, Q., Chen, Q., Iwamoto, Y., Han, X., Chen,
  Y.W., Tong, R., Wu, J.: Semi-supervised segmentation of liver using
  adversarial learning with deep atlas prior. In: MICCAI. pp. 148--156.
  Springer (2019)

\bibitem{zheng2020cartilage}
Zheng, H., Motch~Perrine, S.M., Pitirri, M.K., Kawasaki, K., Wang, C.,
  Richtsmeier, J.T., Chen, D.Z.: Cartilage segmentation in high-resolution {3D
  micro-CT} images via uncertainty-guided self-training with very sparse
  annotation. In: MICCAI. pp. 802--812. Springer (2020)

\end{thebibliography}
\end{document}